\documentclass[letterpaper, 10 pt, journal, twoside]{IEEEtran}

\usepackage{amsmath,amsfonts}
\usepackage{algorithm}
\usepackage{array}
\usepackage[caption=false,font=normalsize,labelfont=sf,textfont=sf]{subfig}
\usepackage{textcomp}
\usepackage{stfloats}
\usepackage{url}
\usepackage{verbatim}
\usepackage{graphicx}
\usepackage{cite}
\usepackage{amsmath}
\usepackage{graphics}  
\usepackage{graphicx}
\usepackage{epsfig} 
\usepackage{cite}
\usepackage{xcolor}
\usepackage{comment}
\usepackage{soul}
\usepackage{adjustbox}
\usepackage{multirow}
\usepackage{wasysym}
\usepackage{colortbl}
\usepackage{hyperref}
\usepackage{algorithm}  
\usepackage{algpseudocode}
\usepackage{amssymb}
\usepackage{nicefrac}
\usepackage[switch]{lineno}
\usepackage{soul}
\usepackage{caption}
\usepackage[font={scriptsize}]{caption}
\hypersetup{ colorlinks = true, citecolor = {green}, linkcolor  = {blue},}
\definecolor{todo}{rgb}{1.0, 0., 0.}

\begin{document}
\title{Anthropomorphic Grasping with Neural Object Shape Completion}

\author{Diego  Hidalgo-Carvajal$^{*, 1,2}$, Hanzhi Chen$^{*, 3}$, Gemma C. Bettelani$^{1}$, Jaesug Jung$^{1}$, Melissa Zavaglia$^{1}$,\\ Laura Busse$^{4}$, Abdeldjallil Naceri$^{1}$, Stefan Leutenegger$^{\ddagger,3}$, Sami Haddadin$^{\ddagger,1,2}$ 
\thanks{Manuscript received: May 25, 2023; Revised: September 01, 2023; Accepted: September, 21, 2023.} 
\thanks{This paper was recommended for publication by Editor Hong Liu upon evaluation of the Associate Editor and Reviewers’ comments. 

This work was funded by the German Research Foundation (DFG, Deutsche Forschungsgemeinschaft) as part of Germany’s Excellence Strategy – EXC 2050/1 – Project ID 390696704 – Cluster of Excellence “Centre for Tactile Internet with Human-in-the-Loop” (CeTI) of Technische Universit{\"a}t Dresden, and by LMUexcellent and TUM AGENDA 2030, funded by the Federal Ministry of Education and Research (BMBF) and the Free State of Bavaria under the Excellence Strategy of the Federal Government and the Länder as well as by the Hightech Agenda Bavaria (ONE MUNICH). This work was also supported by BMBF by funding the project AI.D under the Number 16ME0539K, and by TUM Georg Nemetschek Institute under the project SPAICR. We acknowledge the funding of the Lighthouse Initiative Geriatronics by LongLeif GaPa gGmbH (Project Y), and as part of the SFB~1233 "Robust vision: Inference Principles and Neural Mechanisms", project number 276693517 (TP13).
Please note Sami Haddadin has a potential conflict of interest as a shareholder of Franka Emika GmbH.}  
\thanks{
{* and $\ddagger$ Equal first and last authorships respectively.}}
\thanks{$^{1}$ {\small \tt \{diego.hidalgo-carvajal, djallil.naceri, gemma.bettelani, melissa.zavaglia, jaesug.jung, haddadin\}@tum.de}, 
Technical University of Munich, Germany; TUM School of Computation, Information and Technology (CIT); Chair of Robotics and Systems Intelligence (RSI); Munich Institute of Robotics and Machine Intelligence (MIRMI)}
\thanks{ $^{2}$ also with the Centre for Tactile Internet with Human-in-the-Loop (CeTI)}
\thanks{ $^{3}$ {\small \tt \{hanzhi.chen, stefan.leutenegger\}@tum.de} Technical University of Munich, Germany; TUM School of Computation, Information and Technology (CIT) and MIRMI; Smart Robotics Lab.}
\thanks{ $^{4}$ {\small \tt busse@bio.lmu.de} LMU Munich, Germany; Division of Neuroscience, Faculty of Biology}
\thanks{Digital Object Identifier (DOI): see top of this page.}
}

\markboth{IEEE Robotics and Automation Letters. Preprint Version. Accepted September, 2023}
{Hidalgo-Carvajal \MakeLowercase{\textit{et al.}}: Anthropomorphic Grasping with Shape Completion} 

\maketitle

\begin{abstract}
The progressive prevalence of robots in human-suited environments has given rise to a myriad of object manipulation techniques, in which dexterity plays a paramount role. It is well-established that humans exhibit extraordinary dexterity when handling objects. Such dexterity seems to derive from a robust understanding of object properties (such as weight, size, and shape), as well as a remarkable capacity to interact with them. Hand postures commonly demonstrate the influence of specific regions on objects that need to be grasped, especially when objects are partially visible. In this work, we leverage human-like 
object understanding by reconstructing and completing their full geometry from partial observations, and manipulating them using a 7-DoF anthropomorphic robot hand.
Our approach has significantly improved the grasping success rates of baselines with only partial reconstruction by nearly 30\% and achieved over 150 successful grasps with three different object categories. This demonstrates our approach's consistent ability to predict and execute grasping postures based on the completed object shapes from various directions and positions in real-world scenarios. Our work opens up new possibilities for enhancing robotic applications that require precise grasping and manipulation skills of real-world reconstructed objects.
\end{abstract}


\IEEEpeerreviewmaketitle
\section{INTRODUCTION}
\IEEEPARstart{A}{chieving} human-like dexterous manipulation is a sought-after goal in robotics. Although significant progress has been made to attain this aim, current solutions are limited by both methodologies and hardware constraints. In order to dexterously manipulate an object, two main aspects need to be considered: i) the understanding of the visual object scene, and ii) the grasping strategy. Despite extensive research on these two aspects separately, limited focus has been placed on integrating them into a complete human-like grasping approach.

\begin{figure}[t!]
    \centering
    \includegraphics[width=0.93\columnwidth]{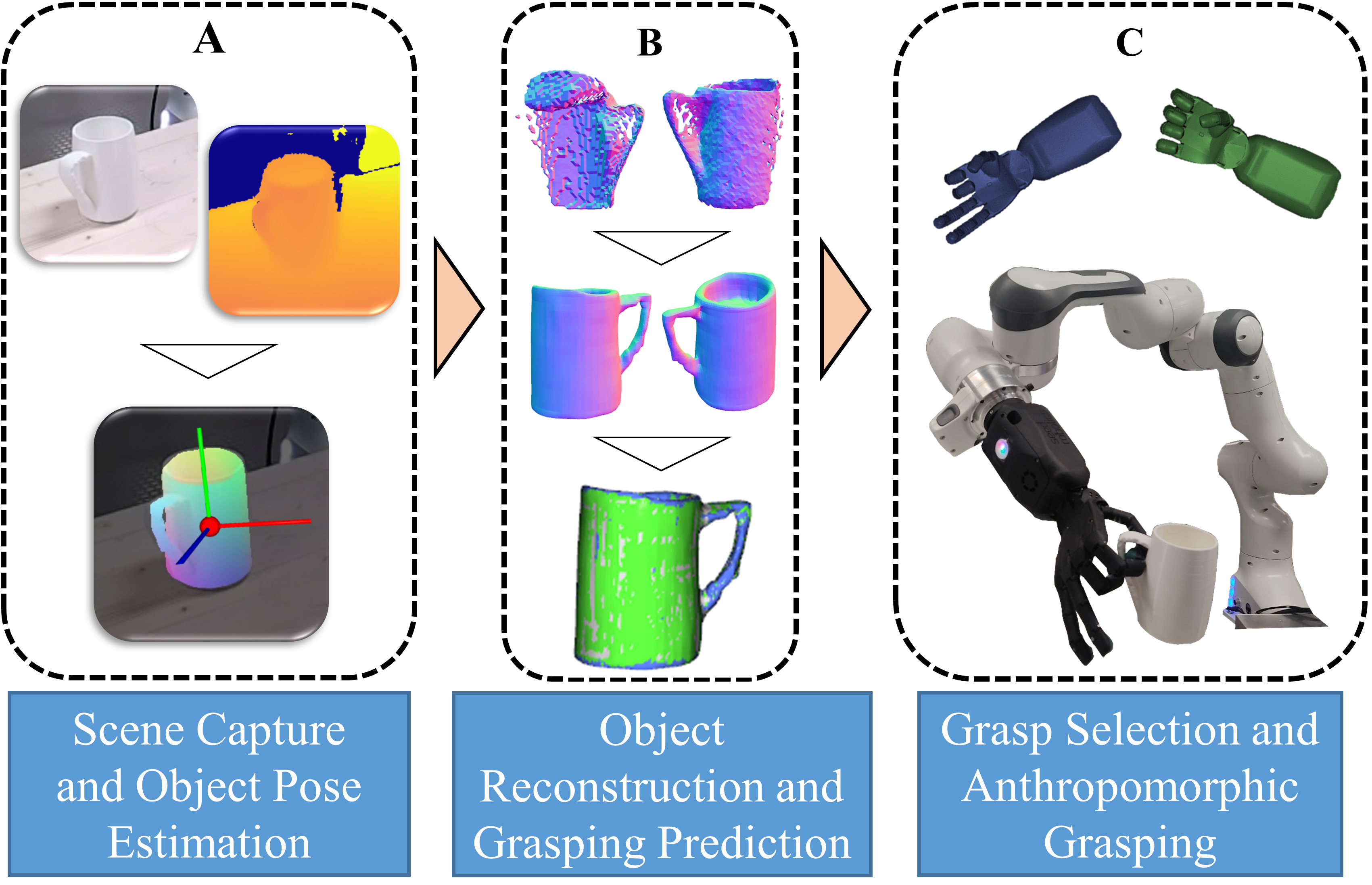}
    \caption{ Overview of our approach. (A) shows sample RGB-D images of the scene and the object pose estimation. (B) shows the object 3D shape reconstruction, completion, and grasping prediction, respectively. (C) shows the possible grasps of the robot hand and the robotic arm-hand system grasping an object in an anthropomorphic manner.
    }
    \label{Fig: Overview}
\end{figure}

In humans, visual object recognition is a complex cognitive process that involves the ability to identify and categorize objects based on their visual features despite substantial ambiguity. It is thought that the brain tackles this challenge with a cascade of \textit{feedforward} processing stages that extract and represent increasingly complex information \cite{herzog2014vision, dicarlo2012Howa}. These computations are supported by \textit{top-down} mechanisms \cite{gilbert2013top}, that can adjust the neural representations based on an internal model of the world derived from prior experience. These mechanisms are often thought to underlie the remarkable ability of humans to robustly recognize objects, even when faced with cluttered environments, ambiguous stimuli, or incomplete information due to partial occlusion \cite{hollingworth2003testing, gilbert2013top}.

Contrary to this, the approaches of the classical artificial vision systems are fundamentally different from human mechanisms. They tend to lead to partial or incomplete object models because they only fuse observable information \cite{bescos2021dynaslam, xu2019mid, runz2018maskfusion}. Recent methodologies have improved those systems by incorporating more human-like approaches by utilizing the complete geometric and semantic information of objects from partial visual information (e.g. \cite{xu2022learning}). While such approaches have performed well in a variety of applications, including robotic grasping, it remains unclear how to leverage them into more end-to-end deep learning  grasping approaches, as such concepts exhibit the potential to leverage learned geometries for more dexterous tasks at hand, where deployment of anthropomorphic robot hands is needed. 

Robotic grasping and manipulation have been extensively studied. However, the majority of studies have focused on non-anthropomorphic simplified hands, such as parallel grippers \cite{mahler2017dexnet}. The reason for this is twofold: i) it is difficult to mimic the human hand motion in an artificial robot hand, and ii) controlling such a convoluted system is an arduous task. In an attempt to find a compromise for these issues, under-actuated tendon-driven robotic hands have been proposed \cite{piazza2019century}. These allow the execution of anthropomorphic manipulation strategies, which can be used to manipulate objects with higher dexterity. Recently, studies have focused on developing generalizable methods to robustly grasp objects from different regions, while understanding the relationship between their geometry and grasping contexts \cite{schmidt2018grasping, Hidalgo_grasping, simeonovdu2021ndf}. 

\begin{figure*}[t!]
    \centering
    \includegraphics[width=0.89\textwidth]{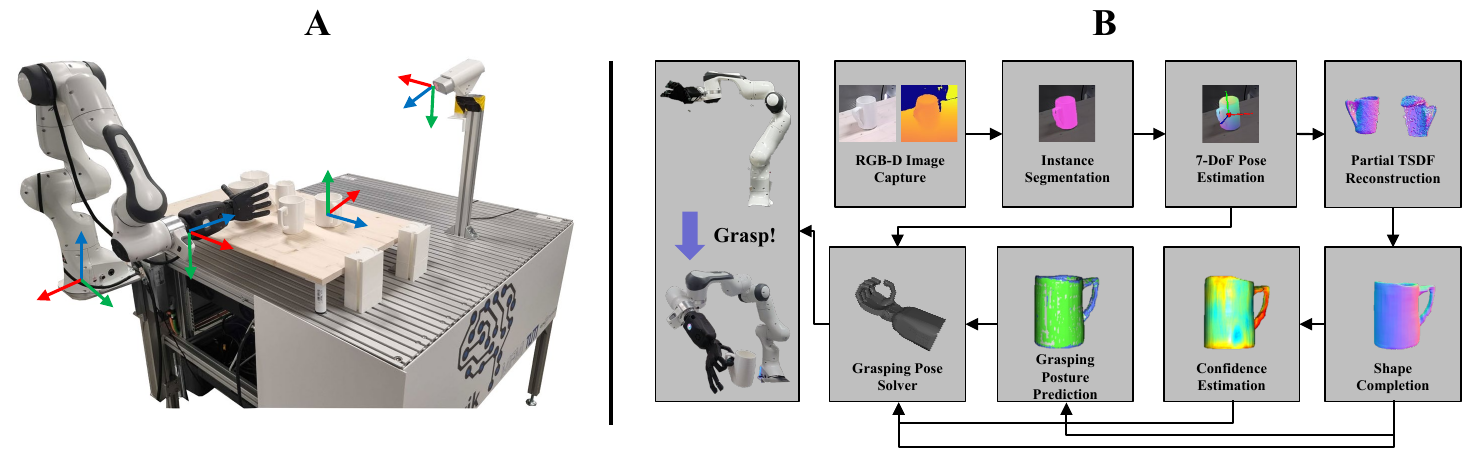}
    \caption{(A) Physical setup of our anthropomorphic hand grasping system. \textcolor{red}{X}\textcolor{green}{Y}\textcolor{blue}{Z}-coordinate systems of the robot base, the robot hand, the object, and the camera are marked. All frames were with respect to the robot base. (B) Overview of our proposed vision-based grasping workflow.}

    \label{Fig: pipeline}
\end{figure*}

In this paper, we proposed a new end-to-end approach (see Fig. \ref{Fig: Overview}) which leverages neural object shape completion to conduct objects grasping in an anthropomorphic manner with a low-cost under-actuated robotic hand.

Our contributions are as follows:
\begin{itemize}
    \item An application in robotic grasping for partial-view conditioned shape completion with shape confidence.
    \item We leverage transferred grasping knowledge from one to multiple instances as proposed in our previous work \cite{Hidalgo_grasping}.
    \item We further design a grasping solver for a 7-DoF robot arm in combination with under-actuated tendon-driven anthropomorphic robot hands.
    \item In a series of real-world experiments, we obtained grasping success rates well above 80\%, close to those achieved with entirely known models and substantially beyond what is achievable solely using (partial) reconstruction.
\end{itemize}
To the best of our knowledge, we are the first to design a system capable of conducting anthropomorphic grasping with object shape completion from single-view visual information in a human-like manner.
\section{RELATED WORKS}

\subsection{Unseen Object's Regions Grasping}
Recent methodologies have provided an alternative solution for vision-based robotic manipulation. Such methodologies are based on categorical semantic consistencies for grasping novel instances. 
On this note, the authors of \cite{wen2021catgrasp} collected a grasping code book in a canonical space and inferred grasp poses through learned dense correspondence. The authors of \cite{simeonovdu2021ndf} proposed to discover 3D point descriptors through shape completion pre-training. Then, they directly transferred grasps to novel instances with matched descriptors.
Work of \cite{wen2022transgrasp} regressed deformation fields for grasp transfer.
The authors of \cite{Wei_2020} focused on a system for fluid human-robot object handovers, where approaching directions were calculated depending on user grasping postures. Notably, although some of these works explored transferable knowledge among categories to conduct grasping on novel instances in a label-efficient manner, they demonstrated such knowledge transfer with parallel-jaw grippers, which have limited dexterity compared to robot anthropomorphic hands. Moreover, unlike our approach, they did not consider the effects of shape and pose variance of unseen objects on transferred knowledge uncertainty. 
Several works have also explored harnessing completed geometry for grasp planning as ours \cite{van2020learning, varley2017shape, jiang2021synergies}, while their reconstruction pipeline often yielded over-smoothed and low-fidelity outcomes, potentially limiting the manipulation dexterity of the robot—such as grasping a mug by its handle.

\subsection{Human-like Dexterous Manipulation}
Grasping strategies are dependent on the dexterity of robot hands and their aptness for specific tasks \cite{fang2020graspnet, mahler2017dexnet, pinto2016supersizing, qin2020s4g, Tieck2020, Gabellieri2020, li2020magichand, valarezo2021natural, rivera2021object, wei2023generalized, ficuciello2019vision, schmidt2018grasping}. In an attempt to find a compromise for the complexity of mimicking human hand motion, and the difficulties of controlling its multiple degrees of freedom (DoF), under-actuated tendon-driven robot hands have been proposed \cite{piazza2019century}. They allow the exertion of anthropomorphic manipulation strategies, which have the potential to expand the current grasping capabilities in robotics. 

A well-established tendency in robotic manipulation is the deployment of grasping strategies for parallel simplified grippers. Works such as \cite{fang2020graspnet, mahler2017dexnet, pinto2016supersizing, qin2020s4g} have predicted stable finger posture placements on objects for such grippers. Nevertheless, their approaches lack generalization for more dexterous end effectors (EE), e.g., anthropomorphic robot hands. On a more recent trend, research efforts have focused on the development of dexterous manipulation using anthropomorphic robot hands \cite{Tieck2020, Gabellieri2020, li2020magichand} emulating human capabilities. Endowing robotic grasping systems with human-like dexterity entails the understanding of human-object interactions. In an attempt to do so, the authors of \cite{GRAB:2020} captured a data set of human-hand-object contacts and developed an anthropomorphic grasping predictor on novel objects, similarly to \cite{Hand_object_contact}. The authors of \cite{valarezo2021natural, rivera2021object} exploited such concepts of human grasping contacts to develop deep learning approaches capable of generating grasps in simulation in a human-like manner using an anthropomorphic robot hand. The authors of \cite{wei2023generalized} followed a similar contact-inspired methodology using a real-world robot-arm-hand system. On a related note, authors of \cite{ficuciello2019vision} proposed a deep learning method, where human-informed policies were used to grasp different objects with different anthropomorphic grasps. The authors of \cite{schmidt2018grasping} went further and used depth images and deep convolutional neural networks to grasp objects using their entire geometry, without the limitation of using parallel grippers.
The aforementioned approaches were able to grasp objects stably in simulation and real world experiments. Nevertheless, in most cases, they used simplified grippers, and could not choose grasping regions or approaching directions to the objects. Additionally, they did not analyze relations between different grasping types and the entirety of the objects geometry. This may be a limiting factor when manipulating objects in a dexterous manner, as different object regions may require different grips depending on the task at hand.

\section{METHODS}
As shown in Fig. \ref{Fig: Overview}, our proposed approach started with the capture of a static RGB-D image of the grasping scene. This image was used as an input to estimate present object shapes. The parts of the objects that were not visible to the camera were completed using our object shape reconstruction pipeline inspired by \cite{xu2022learning}. Specifically, we focused on drinking cups, bottles, and bowls. Our grasping posture predictor, proposed in \cite{Hidalgo_grasping}, was used to infer grasping postures in the entire geometry of objects after shape completion. Finally, our robot hand-arm system grasped the objects in areas of interest with predefined anthropomorphic grasping sequences using our proposed hand grasping posture selection solver. The following sections detail each component of our proposed approach.

\subsection{System Setup and Scene Capture}
\label{sec: setup}
We used a robot hand-arm system, consisting 
of a 7-DoF Panda robot (Panda + FCI Lizenz, Franka Emika, Germany) and a 7-DoF (2 for the thumb, 1 for the index, 1 for the middle, 1 for both ring and little fingers, and 2 for the wrist) anthropomorphic robot hand (RH8D, Seed Robotics, Portugal). Our setup consisted of a fixed metallic table of dimensions $1192 \times 1100 \times 860\text{ mm}$, with an attached secondary metallic base $73\text{ mm}$ beneath. The secondary metallic base had dimensions $226 \times 190\text{ mm}$ and supported the Franka Panda robot. The main metallic table was located on the second octant of the robot's workspace $(-X,+Y,+Z)$. For our experiments, we placed a wooden table of dimensions $800 \times 500\text{ mm}$ on top of the metallic table ($120\text{ mm}$ above), where objects were placed 
during our experiments. Fig. \ref{Fig: pipeline}-A shows our robot setup.

To capture the grasping scene, we used a programmable multi-mode RGB-D camera (Azure Kinect DK, Microsoft, USA), which was mounted on a fixed platform using a customized 2-DoF assembly, allowing pitch and yaw adjustment (see Fig. \ref{Fig: pipeline}-A). The camera was located at a height of $515\text{ mm}$ with respect to the metallic table, and was tilted $30^{\circ}$ downwards pointing at the reachable workspace of the robot arm in our setup (second octant of its workspace: $-X,+Y,+Z$). 

\subsection{Object Pose Estimation and Shape Reconstruction}

\label{sec: meshcomplete}
In this part, our goal was to retrieve geometric information of a novel instance from a pre-defined category. We used an object detector, a pose estimator, and a shape mapper parameterized by three individual deep neural networks, denoted as $f_\text{detect}(\cdot)$, $f_\text{pose}(\cdot)$, and $f_\text{map}(\cdot)$, respectively. Given one single-view RGB-D frame input $[\mathbf{I}, \mathbf{D}]$, we first used the object detector \cite{kirillov2020pointrend} to acquire the foreground mask of the object of interest with $\mathbf{S}=f_\text{detect}(\mathbf{I})$. Note $\mathbf{I}, \mathbf{D}, \mathbf{S}$ shared the same resolution ($\mathbf{I}, \mathbf{D}, \mathbf{S} \in \mathbb{R}^{1536 \times 2048}$)
Then we acquired foreground point clouds using the segmented depth $\mathbf{D}_{\text{obj}}=\mathbf{S}\odot\mathbf{D}$, its corresponding pixel coordinates $\mathbf{u}_{\text{obj}}$ and camera intrinsics matrix $\mathbf{K}$ $\in \mathbb{R}^{3\times3}$ with $\mathbf{x}_{\text{obj}}=\pi^{-1}(\mathbf{D}_{\text{obj}}, \mathbf{u}_{\text{obj}}, \mathbf{K})$, and passed $\mathbf{x}_{\text{obj}}$ to the pose estimator~\cite{li2023generative} to acquire the objects' 7-DoF pose (rotation, translation, and scale) between its pre-defined canonical frame and camera frame: $ [\mathbf{R}, \mathbf{t}, s]=f_\text{pose}(\mathbf{x}_{\text{obj}})$, where $\mathbf{R} \in \text{SO}(3)$, $\mathbf{t} \in \mathbb{R}^3$, $s \in \mathbb{R}$.

To acquire the complete shape regardless of occlusions, we based the shape mapper from \cite{xu2022learning}, while adding gradients regularization for the occupancy of free space during training and removed heavy test-time shape optimization during inference for the sake of speed. We transformed the observed point clouds to the canonical frame with $\mathbf{x}_{\text{cano}}=s\mathbf{R}\mathbf{x}_{\text{obj}}+\mathbf{t}$ and further acquired a partial truncated signed distance function (TSDF) volume $\mathbf{V}_{\text{cano}}$  $\in \mathbb{R}^{64\times64\times64}$ by voxelizing $\mathbf{x}_{\text{cano}}$. Conditioning on $\mathbf{V}_{\text{cano}}$, the shape mapper predicted the complete shape represented by voxelized occupancy probability with $\mathbf{P}_{\text{cano}}=f_{\text{map}}(\mathbf{V}_{\text{cano}})$. Finally, we extracted the complete mesh from the voxel grids using the multi-resolution iso-surface extraction strategy from \cite{mescheder2019occupancy}. We further inferred shape confidence for each mesh vertex $\mathbf{v}$ $\in \mathbb{R}^3$ represented by the norm of its gradient w.r.t. the occupancy probability ($||\partial \mathbf{P}_{\text{cano}}[\mathbf{v}]/\partial \mathbf{v}||_2$), which were further used as one of the criteria to select reliable grasps introduced in Section~\ref{sec: anthropomorphic_hand_grasping}.

\subsection{Grasping Posture Prediction}

\label{sec: graspposture}

Following the insights for grasping transferability in \cite{Hidalgo_grasping}, we adopted our previously proposed grasping posture predictor to select plausible grasping postures for our robot arm-hand system to approach and grasp objects. The grasping posture predictor was also parameterized by a deep neural network denoted as $f_\text{posture}(\cdot)$. For each queried vertex $\mathbf{v}$ from the complete mesh provided by Section \ref{sec: meshcomplete}, it predicted a grasping posture label $g$ corresponding to either a medium wrap grasp (MW), tripod grasp (T), or non-graspable regions (NG) defined in \cite{feix2015grasp} with $g=f_\text{posture}(\mathbf{v})$. 

\subsection{Hand grasping posture selection solver}
\label{sec: anthropomorphic_hand_grasping}
\begin{figure}[t!]
    \centering
    \includegraphics[width=1\columnwidth]{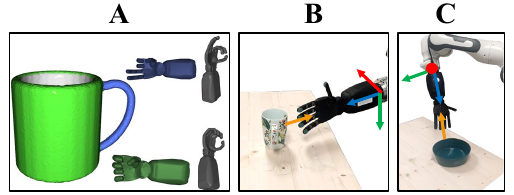}
    \caption{(A) Object-centric grasping posture labeling. Green indicates medium wrap, and blue indicates tripod grasps. (B) Side-approaching grasp. (C) Top-approaching grasp. We marked the \textcolor{orange}{normal}
    of the selected vertex and \textcolor{red}{X}\textcolor{green}{Y}\textcolor{blue}{Z} indicates the robot hand frame.}
    \label{Fig: Objects_and_labels}
\end{figure}

We used the setup mentioned in Section \ref{sec: setup} to exert anthropomorphic grasps with different orientations in different areas of objects of interest. Once the objects had assigned grasping posture predictions for their entire geometry, we selected the vertices of interest for grasping. Due to the constraints of our setup, we chose vertices whose normal vectors pointed on the first octant of the robot base frame $(+X, +Y, +Z)$. We then selected candidate grasping points based on two criteria, namely, according to the highest shape confidence, and in an arbitrary manner, as an alternative when confidence values were not available (see Section \ref{sec: experiments}). We then used a threshold of $45^{\circ}$ on the inclination of the vertex's normal with respect to the XY plane of the robot's frame, to select either a side-approaching (Fig. \ref{Fig: Objects_and_labels}-B) or top-approaching grasping (Fig. \ref{Fig: Objects_and_labels}-C). 
For side-approaching grasps (e.g. for bottles, or drinking cups being grasped from the side): we used vertices whose $Z >= 45\text{ mm}$ above the table. This safety distance (from the center of the robot hand palm to its ulnar end along its coronal plane \footnote{Terminology in regard to the robot hand corresponds to the anatomical conventions for human hands.}) guaranteed that the center of the robot hand's palm could approach the chosen grasping point without colliding with the table.
We used top-approaching grasps (e.g. for bowls, or drinking cups being grasped from above) for vertices whose normal formed an angle of $>= 45^{\circ}$ with the table. Normally normal vectors with such inclination were located at the top of the objects. In such instances a side-grasping approach was unfeasible.

\begin{figure}[t!]
    \centering  \includegraphics[width=1\columnwidth]{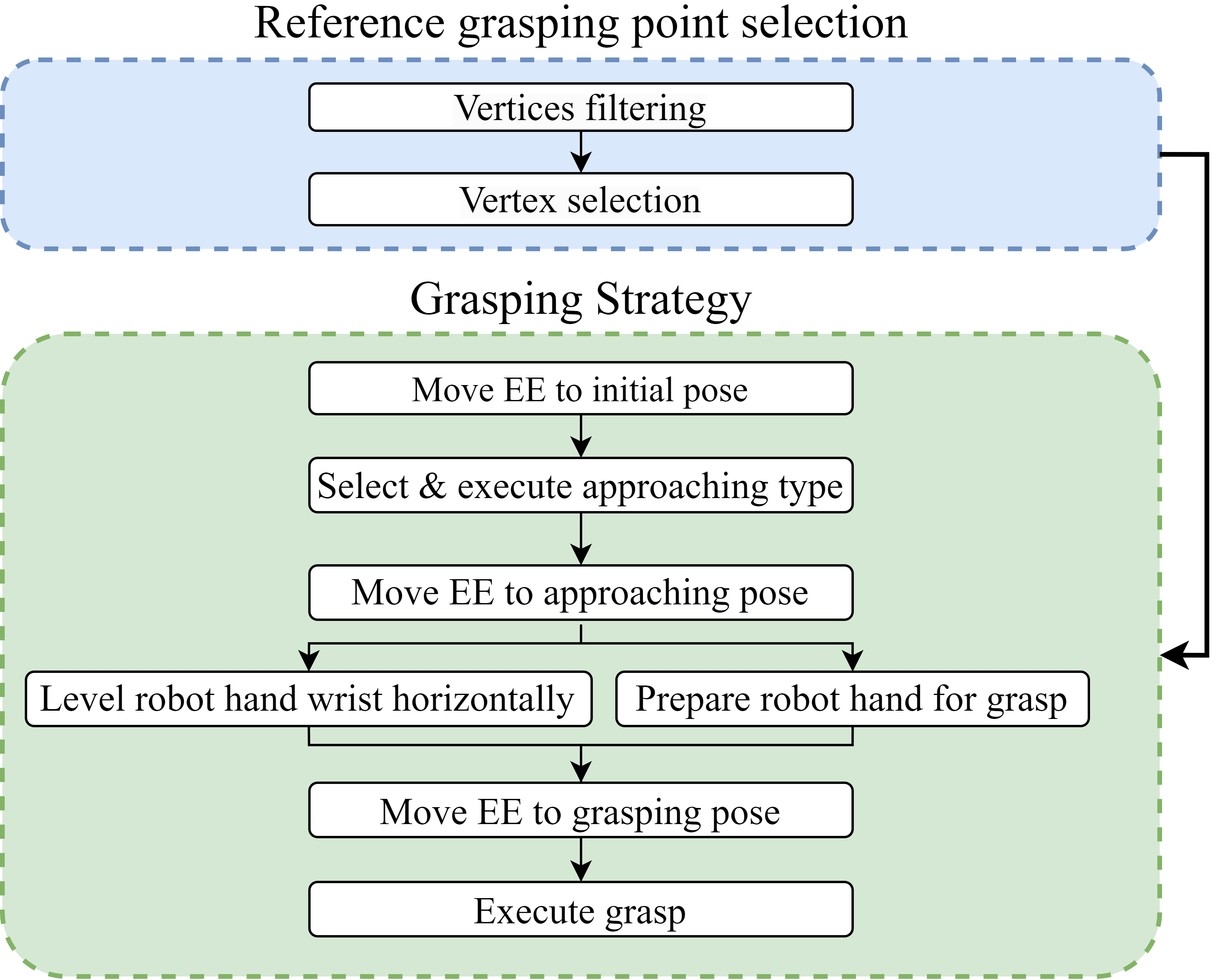}
    \caption{Hand grasping posture selection solver. See Section \ref{sec: anthropomorphic_hand_grasping} for details.}
    \label{Fig: Grasping_flowchart}
\end{figure}

Our grasping strategy consisted of the following steps (see Fig. \ref{Fig: Grasping_flowchart}): 
\begin{enumerate}
    \item First, the robot arm-hand system moved to an idle initial position at the top of the manipulation workspace visible by the camera.
    \item Then we calculated a suitable robot hand wrist frame so that its axis perpendicular to the hand transverse plane pointing in the distal direction was opposite to the normal vector 
    of the selected grasping vertex (see Fig. \ref{Fig: Objects_and_labels}). The other two axes on the wrist frame were calculated so the $Z$-axis of the robot base frame lay on the coronal plane of the robot hand (in flat configuration) for side-approaching grasps, and on the sagittal plane for top-approaching grasps, respectively. All frames and vectors were expressed in the robot base frame.    
    \item Following this, the robot's EE (tips of the middle and ring fingers of the robot hand in a flat configuration) was moved to an approaching position. This position was calculated as the point where the tips of the middle and ring finger 
    were at a distance of $50\text{ mm}$ to the vertex of interest along its corresponding normal vector.
    \item Subsequently, we prepared the robot hand for grasping as follows. 
    We developed an algorithm to keep either the sagittal plane (side-approaching grasp) or the coronal plane (top-approaching grasp) of the robot hand parallel to the $XY$ plane of the robot frame (horizontal leveling).
    This was achieved by abducting (side-approaching) or extending (top-approaching) the wrist an angle $\gamma$ (angle between the normal vector of the chosen grasping point and the horizontal $XY$ plane). Since leveling the hand horizontally moved the hand upwards, we added compensation distances, so the robot hand's palm center could reach the selected grasping vertex on the object. These distances were calculated as shown in equations \ref{eq: compensated_distances_x}, \ref{eq: compensated_distances_y}. 
    The distance $d_{{e}_{x'}}$ was computed as:
    \begin{equation}\label{eq: compensated_distances_x}
      d_{{e}_{x'}} = l (1-\cos(\gamma)),
    \end{equation}
    where $l$ is the distance from the robot hand's wrist  
    to the tip of the middle finger (EE) when the hand is in a flat configuration. 
    $d_{{e}_{x'}}$ was compensated along the $\mathbf{e}_{x'}$ axis
    \begin{equation}\label{eq: compensated_axes_x}
      \mathbf{e}_{x'} = \left(\left[ \begin{matrix} 0\:0\:1 \end{matrix} \right] \times \mathbf{n}\right) \times \left[ \begin{matrix} 0\:0\:1 \end{matrix} \right],
    \end{equation}
    where $\mathbf{n}$ is the axis opposing the normal vector to the selected grasping vertex (see Fig. \ref{Fig: Objects_and_labels}).
    The distance $d_z$ was compensated along the $[0, 0, -1]$ axis:
    \begin{equation}\label{eq: compensated_distances_y}
      d_z = l \sin(\gamma).
    \end{equation}        
    Finally, these compensation distances were combined into the compensation displacement vector $\mathbf{c}$ and added to the end effector's position. For side-approaching grasps, we extended the hand's wrist $45^{\circ}$ (its maximum extension angle) and fully abducted the thumb in the palmar direction.
    \item Afterwards, the EE was moved a distance of $150\text{ mm}$ along the direction opposed to the normal vector at the vertex of interest ($\mathbf{n}$). This distance was set as the sum of the aforementioned approaching distance of $50\text{ mm}$ and the distance from the center of the palm to the tip of the middle finger or EE ($100\text{ mm}$). The aforementioned displacement compensation ($\mathbf{c}$) guaranteed that the palm of the hand was positioned next to the vertex of interest during grasping.
    \item Finally, a robot hand pre-programmed finger grasping sequence, corresponding to either a medium wrap or a tripod grasp \cite{feix2015grasp} was executed. These grasping sequences were independent of the aforementioned approaching types. Once the hand was closed, the object was lifted vertically (a distance of $100\text{ mm}$ for side-approaching, and $200\text{ mm}$ for the top-approaching grasps). We used the motor currents on the robot hand to detect successful grasping. A successful medium wrap and tripod grasp were considered when the motor currents of at least four and three motors were higher than $400\text{ mA}$ for $\geq4\text{ s}$, respectively. 
\end{enumerate}

\subsection{Network Training Protocol}

We used pre-trained off-the-shelf network weights for the object detector and pose estimator, provided by \cite{wu2019detectron2} and \cite{li2023generative}, respectively. For the shape mapper for completion, we used 3D models provided by ShapeNet repository \cite{chang2015shapenet} to first render single-view depth train data using BlenderProc2 \cite{Denninger2023}. Then we adopted the same loss function as in \cite{xu2022learning} and trained the network for 20K iterations with a batch size of 48. 

For our grasping predictor,  we manually labeled one known object instance with the three grasping posture labels introduced in Section \ref{sec: graspposture}. The known 
objects were represented by a triangular mesh 
with an average triangle side length of 2.5 mm. We followed the same training protocol as \cite{Hidalgo_grasping}.

\subsection{Experiments}
\label{sec: experiments}

Our experiments consisted of three parts. We tested our grasping framework using five 3D-printed drinking cups in the first part. The second part was done with five arbitrarily selected real-world drinking cups. We conducted grasping experiments for five bottles and five bowls in the third part. We calibrated the camera of our setup before the experiments using a modified version of the calibration pipeline proposed in\footnote{https://github.com/marcoesposito1988/easy\textunderscore handeye\textunderscore demo}. The camera calibration was done with respect to the robot base frame and it allowed the direct representation of the reconstructed meshes in the robot base frame.

\subsubsection{Grasping 3D-printed drinking cups}
\label{sec: grasp_3d_mugs}

The experiments in this section were threefold. For each part, the 3D-printed drinking cups were placed on top of a wooden plate (simulating a regular table) in the camera's visible area. First, we tested our grasping strategy using the fully known mesh models of the objects (available to us, as we 3D printed them). We then arbitrarily selected points whose coordinates and normal vectors satisfied our setup grasping constraints, and would not result in a singular configuration or a collision. This was done because no object completion was required and, therefore, completion confidence values were not available. We then executed our grasping routine. We categorized a grasping attempt as a success if the hand was able to hold the object for four seconds after it was lifted. In the second part, we took a single static RGB-D image of the scene and passed it to reconstruct the complete shapes of the desired drinking cup as introduced in Section \ref{sec: meshcomplete}. We then grasped the drinking cups using the points with the highest shape confidence. Finally, we repeated the second part of the experiments with the object 3D reconstruction but without completion. We selected points randomly from the ones satisfying workspace constraints (see Section \ref{sec: anthropomorphic_hand_grasping}).
\begin{figure}[t!]
    \centering
    \includegraphics[width=1\columnwidth]{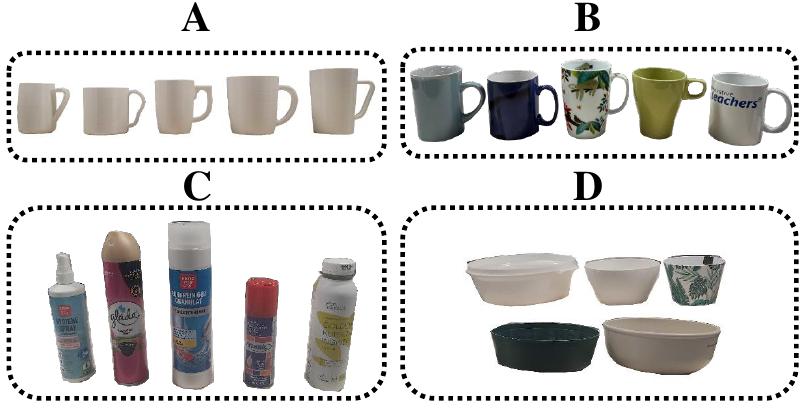}
    \caption{Tested objects in grasping experiments. (A) 3D-printed drinking cups. (B) Real 
    drinking cups. (C) Real bottles. (D) Real bowls}
    \label{Fig: Drinking_cups}

\end{figure}
\subsubsection{Grasping real-world drinking cups}
\label{sec: grasp_real_mugs}
In this part of the experiments, we recreated the second and third parts of the experiments in section \ref{sec: grasp_3d_mugs} using five real-world drinking cups. We first grasped drinking cups which were reconstructed and completed using color and depth information from a static image taken from the camera. We chose grasping points according to the highest shape confidence values and executed our grasping procedure. We then performed our grasping strategy on the same objects and positions, but without 3D shape completion, using points that satisfied the setup constraints.

\subsubsection{Grasping real-world bottles and bowls}
\label{sec: grasp_real_bottles_bowls}
In these experiment trials, we followed the approach of the first part of experiments in section \ref{sec: grasp_real_mugs} and applied it to two additional object categories. We grasped bottles and bowls, whose shape was reconstructed and completed, choosing points according to the highest shape confidence values, and executed our grasping procedure. We tested five different instances for each category.

\section{RESULTS}

\begin{figure*}[t!]
    \centering
    \includegraphics[width=0.89\textwidth]{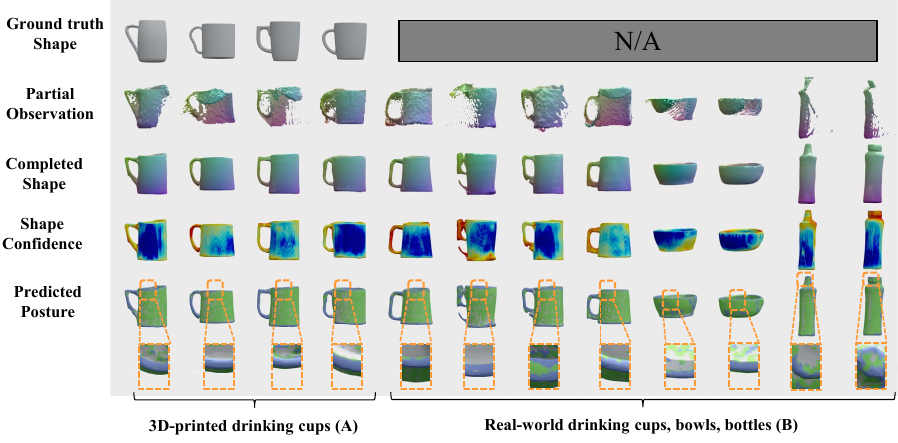}
    \caption{Ground truth meshes, partial observations fed to the object reconstruction pipeline, and its outputted completed shapes, as well as their shape confidence (blue indicates high confidence, red indicates low confidence). The fifth row shows grasping predictions on the completed shapes (medium wrap in green, and tripod in blue). The last row below (in orange boxes) depicts a zoomed-in view on object rims for top-approaching grasps.}
    \label{Fig: qualitative}
\end{figure*}

\subsection{Shape accuracy of 3D-printed drinking cups}
We first used the five 3D-printed drinking cups with ground-truth meshes to showcase the effectiveness of our object shape completion pipeline by comparing the shape accuracy of the extracted mesh after completion and the raw mesh from partial observations. The metrics for shape accuracy evaluation, chamfer distance L1, completeness, and normal consistency were first proposed by \cite{mescheder2019occupancy}, and we kindly refer interested readers to it for detailed definitions. 
For chamfer distance (L1), we significantly improved the accuracy of raw and incomplete geometry from 0.0397 to 0.0157 with our completion pipeline, such gain was also reflected in completeness (from 0.0597 to 0.0124) and normal consistency (from 0.6842 to 0.8969). 
These results are quantitatively and qualitatively demonstrated in Table \ref{tab: results_shape}, and in Fig. \ref{Fig: qualitative}, respectively.

\subsection{Grasping 3D-printed drinking cups}
For the first phase of our grasping experiments, we grasped five 3D-printed drinking cups 33 times in three stages, totaling 99 times. In each of the stages, we grasped the cups with a medium wrap (18 times) and tripod (15 times) grip type. In the first stage of the experiments, we grasped the drinking cups using their known meshes. We achieved a success rate of 87.87\%, grasping the cups with a medium wrap 16 times (88.89\%) and with a tripod grip 13 times (86.67\%)
. In the second stage, we reconstructed and completed the 3D shape of the objects and were able to grasp them successfully at a rate of 81.82\%, with 17 successful grasps using the medium wrap (94.44\%) and 10 tripod grips (66.67\%). In the third stage, we grasped the objects only with the 3D shape reconstruction but without completion. The success rate for this part of the experiments was 54.6\%, being able to grasp the objects 8 times with the medium grap (44.44\%) and 10 times with the tripod grip (66.67\%). Table \ref{tab: results_success} shows the success results 
of the 3D printed objects.

\subsection{Grasping real-world drinking cups}
In the second part of our grasping experiments, we grasped five real-world drinking cups in two stages, 30 times each. In each one of the stages, we grasped the objects 15 times with the medium grasp and 15 times with the tripod grips. When we used reconstruction and completion, we achieved a success rate of 83.33\%, being able to successfully grasp drinking cups with a medium wrap grip with a success rate of 73.33\%. When using the tripod grip, the success rate increased to 93.33\% (14 times). However, when using non-completed 3D models the overall accuracy was 50.00\%. We grasped drinking cups successfully with the medium wrap and tripod grips 46.67\% and 53.33\% of the time, respectively.
\subsection{Grasping real-world bottles and bowls} 
Following the experimental procedure of section \ref{sec: grasp_real_bottles_bowls}, we conducted grasping experiments for 10 additional objects, namely, 5 commercial bottles, and 5 commercial bowls. We grasped each object 10 times, totaling 100 experiments. For each trial, we executed the grasping posture suggested by our grasping posture predictor (i.e. Medium Wrap or Tripod). Our framework reached successful grasps in 86.00\% of the instances for bottles and 82.00\% of the instances for bowls. 
\begin{table}[t!]
\centering
\caption{Shape accuracy of 3D-printed drinking cups (Chamfer distance L1, Completeness and normal consistency) $\downarrow$: lower is better, $\uparrow$: higher is better.} 
\begin{adjustbox}{width=0.66\columnwidth,center}
\begin{tabular}{|l||*{2}{c|}}\hline

\multicolumn{1}{|l||}{\multirow{3}{*}{\begin{tabular}{@{}c@{}}Shape \\ Accuracy\end{tabular}}} & \multicolumn{2}{c|}{3D-printed drinking cups} 
\\ [0.5ex] 

\cline{2-3}
& Reconstruction & Reconstruction \\
& and completion & only \\
\hline\hline
Chamfer Dist. (L1) ($\downarrow$)   & \textbf{0.0157} & 0.0397 \\ 
\cline{2-3}
Completeness  ($\downarrow$)        & \textbf{0.0124} & 0.0597 \\   
\cline{2-3}
Normal Consis. ($\uparrow$)         & \textbf{0.8969} & 0.6842  \\ 

\hline
\end{tabular}
\end{adjustbox}
\label{tab: results_shape}
\end{table}

\begin{table}[t!]
\centering

\caption{Success rates (\%) for experiments with 3D printed and real world drinking cups. MW = Medium Wrap, T = Tripod.}
\begin{adjustbox}{width=0.9\columnwidth,center}
\begin{tabular}{|l||*{5}{c|}}\hline

\multicolumn{1}{|l||}{\multirow{3}{*}{\begin{tabular}{@{}c@{}}Grasping \\ Type\end{tabular}}} & \multicolumn{3}{c|}{3D-printed drinking cups} & \multicolumn{2}{c|}{Real-world drinking cups}\\ [0.5ex]
\cline{2-6}
&   & \multicolumn{4}{c|}{Unknown models}   \\

\cline{3-6}

& Known & Reconstruction & Reconstruction & Reconstruction & Reconstruction\\
& models & and completion & only & and completion & only\\

\hline\hline
MW  & 88.89 & \textbf{94.44} & 44.44 & \textbf{73.33} & 46.67\\
\cline{2-6}
T & 86.67 & \textbf{66.67} & \textbf{66.67} & \textbf{93.33} & 53.33\\
\cline{2-6}
Total & 87.87 & \textbf{81.82} & 54.60 & \textbf{83.33} & 50.00\\

\hline
\end{tabular}
\end{adjustbox}
\label{tab: results_success}

\end{table}

\section{DISCUSSION}

In this work, we proposed a novel approach that combines object shape completion with a grasping strategy. Our approach can be robustly used to grasp partially visible objects in an anthropomorphic manner. The proposed approach benefits from the anthropomorphism of a multi-fingered robot hand, resembling more closely human grasping capabilities and strategies. Additionally, our pipeline can be run automatically. 

Our object shape reconstruction method demonstrated its strong ability to retrieve smooth, clean, and complete geometric information from raw, noisy, and partial observations, potentially caused by sensed depth noise, imperfect segmentation, or occlusions. To illustrate this, Fig. \ref{Fig: qualitative} showed that partial reconstruction of the handles was deficient compared to other parts of the drinking cups. This is due to their thin structures and high curvatures. Nevertheless, our reconstruction pipeline was still able to retrieve fine details of the geometry even under such imperfect conditions. The estimated shape confidence for the handle regions was low. This is because the conditioned input in these regions was prone to noise. On a similar note, although the drinking cup handles were not visible due to object self-occlusions (see column 6 in Fig. \ref{Fig: qualitative}), our method was still able to estimate handle shapes reasonably. Nevertheless, such estimations had higher uncertainty, as expected. Regarding the body part of the cups, even though the completed regions had relatively low confidence values compared to those of visible regions (see columns 1 and 2 in Fig. \ref{Fig: qualitative}), several regions still exhibited high confidence values. This is due to the "symmetry prior" learned from our shape mapper network through 3D completion training. We further demonstrated the results of bowls and bottles in columns 9-12, which again yielded promising full-shape reconstructions.

The integration of our shape reconstruction and completion pipeline along with our hand grasping posture selection solver allowed us to robustly grasp different types of objects in spite of significant object shape occlusions. Our approach brings dexterity a step forward by closing the gaps on missing components of previous works \cite{schmidt2018grasping, simeonovdu2021ndf}, where no full object anthropomorphic grasp predictions were analyzed nor exerted. Although we were able to grasp objects using only partially reconstructed meshes from the RGB-D images, the robustness ($\sim 50\%$) of the grasps depended on the quality of the reconstruction, which in turn depended on the position of the objects and on the grasping points selection. Additionally, since the camera was placed on the opposite side of the robot's grasping space, the reconstructed geometries on the robot's grasping space had limited feasible regions for grasping. However, when using 3D shape completion, we were able to select the grasping regions on the robot's grasping space according to two different criteria (highest completion certainty and arbitrarily, as described in Section \ref{sec: grasp_3d_mugs}) and grasp the drinking cups with a success rate up to 83.33\% using both medium wrap and tripod grips in both 3D printed and regular cups. As expected, the success rate was lower compared to when we used the fully known object models (87.87\%), as explained in Section \ref{sec: grasp_3d_mugs}. The high success rate for other categories (86\% for bottles and 82\% for bowls) further verified the scalability potential of our pipeline. 

Our physical setup and our right-handed grasping strategy limited our graspable vertices, as explained in section \ref{sec: anthropomorphic_hand_grasping}. This vertices selection can be adjusted effortlessly for setups with objects located in other octants of the robot's workspace. Our framework has not been tested in daily real-world scenarios, for instance highly cluttered scenes at home. Nevertheless, the generalization capabilities of our approach show potential as a real-world pipeline capable of achieving noteworthy results.

The grasping types were selected based on our grasping predictor proposed in \cite{Hidalgo_grasping}. Due to the geometric variance of unseen objects, some regions of the objects could be incorrectly labeled as non-graspable regions, or with an outlier grasping type, which was unfeasible to accomplish (\textit{c.f.} "Predicted Posture" in Fig. \ref{Fig: qualitative}). Our pipeline mitigated this issue by carefully designing the posture selection solver considering geometric features and shape confidence of the objects, to select reliable approach points for the robot hand. In regard to the postures we used in this work, we selected objects that required the exertion of a medium wrap and tripod grips, as shown in \cite{Hidalgo_grasping}. This selection allowed the demonstration of power and precision grasps, which are crucial for human anthropomorphic grasping. These two grasping postures, additionally, represent motions along the first two synergies proposed by \cite{santello1998postural}, and are among the five most common postures in activities of daily living \cite{feix2014analysis}. Since robot hand grasps are treated as an independent module in our framework, additional postures, using more sophisticated robot hands can be implemented. With our approach, we attempt to show that exact contact points or grasping forces are not required for high grasping success rates, emulating human knowledge and manipulation strategies into robotic grasping. Nevertheless, our framework's modularity can be exploited by including robot hands that allow customization of further parameters during grasps such as finger forces.

Since our grasping predictor and shape estimator played a major role in our pipeline, improvements in their accuracy would lead to higher success rates. This may be achieved by adding more category-specific training samples. However, this could bring undesirable inefficiency to the preparation process as annotating grasping postures, for instance, is labor-intensive as discussed in \cite{Hidalgo_grasping}. It is hence worthwhile to explore the trade-off between the training data size and grasping performance required for general grasping frameworks as ours.
\section{CONCLUSION}

In this work, we successfully implemented an end-to-end approach (see Fig. \ref{Fig: Overview}) that integrates object shape completion, grasping posture prediction, and a robot arm-hand anthropomorphic grasping strategy to grasp unseen objects in a real-world setup. Our approach can automatically reconstruct and infer complete 3D models of novel objects from a single static view under occlusions. We can then predict grasping postures associated with the entire geometry of the objects in a highly label-efficient manner. Our proposed grasping strategy, shown in Fig. \ref{Fig: Grasping_flowchart}, allowed us to robustly grasp and lift a variety of objects over 150 times using a low-cost anthropomorphic robot hand. Given our method's robustness against object occlusions and its aptness to grasp objects in arbitrary regions from arbitrary directions in an anthropomorphic manner, the proposed approach shows high potential for applications where assistive robots play a crucial role, for example, scenarios of activities of daily living for elderly care. Our framework's current state requires the selection of objects that fit within the used robot hand workspace. This might limit the object categories to be grasped, similarly to a human hand. Additionally, the framework does not operate with low-latency constraints. Nevertheless, it has the potential to be deployed with hardware acceleration modules. For future work, we aim to integrate uncertainty estimation for the grasping posture predictor and extend our pipeline to be category-agnostic so as to handle more previously unseen objects.

\bibliographystyle{IEEEtran}
\typeout{}
\bibliography{bibliography}
\end{document}